\documentclass[sn-mathphys]{sn-jnl}

\usepackage{MnSymbol}

\jyear{2021}%

\theoremstyle{thmstyleone}%
\theoremstyle{thmstyletwo}%

\theoremstyle{thmstylethree}%

\begin{document}

\title[Article Title]{EVA2.0: Investigating Open-Domain Chinese Dialogue Systems with Large-Scale Pre-Training}


\author[1,2]{\fnm{Yuxian} \sur{Gu}}
\equalcont{These authors contributed equally to this work.}

\author[1,2]{\fnm{Jiaxin} \sur{Wen}}
\equalcont{These authors contributed equally to this work.}

\author[1,2]{\fnm{Hao} \sur{Sun}}
\equalcont{These authors contributed equally to this work.}

\author[1,2]{\fnm{Yi} \sur{Song}}

\author[1,2]{\fnm{Pei} \sur{Ke}}

\author[1,2]{\fnm{Chujie} \sur{Zheng}}

\author[1,2]{\fnm{Zheng} \sur{Zhang}}

\author[2]{\fnm{Jianzhu} \sur{Yao}}

\author[3]{\fnm{Lei} \sur{Liu}}

\author[1,2]{\fnm{Xiaoyan} \sur{Zhu}}

\author[1,2]{\fnm{Minlie} \sur{Huang}}

\affil[1]{The CoAI group, Tsinghua University, Beijing, China}

\affil[2]{Department of Computer Science and Technology, Tsinghua University, Beijing, China}

\affil[3]{Department of Electrical Engineering and Computer Science, York University}



\abstract{Large-scale pre-training has shown remarkable performance in building open-domain dialogue systems. However, previous works mainly focus on showing and evaluating the conversational performance of the released dialogue model, ignoring the discussion of some key factors towards a powerful human-like chatbot, especially in Chinese scenarios. In this paper, we conduct extensive experiments to investigate these under-explored factors, including data quality control, model architecture designs, training approaches, and decoding strategies. We propose EVA2.0, a large-scale pre-trained open-domain Chinese dialogue model with 2.8 billion parameters, and will make our models and codes publicly available. Automatic and human evaluations show that EVA2.0 significantly outperforms other open-source counterparts. We also discuss the limitations of this work by presenting some failure cases and pose some future research directions on large-scale Chinese open-domain dialogue systems.}

\keywords{Natural language processing, deep learning, large-scale pre-training, dialogue systems, Chinese open-domain conversational model.}



\maketitle

\section{Introduction}

\begin{figure}[t]
  \centering
  \includegraphics[width=0.7\linewidth]{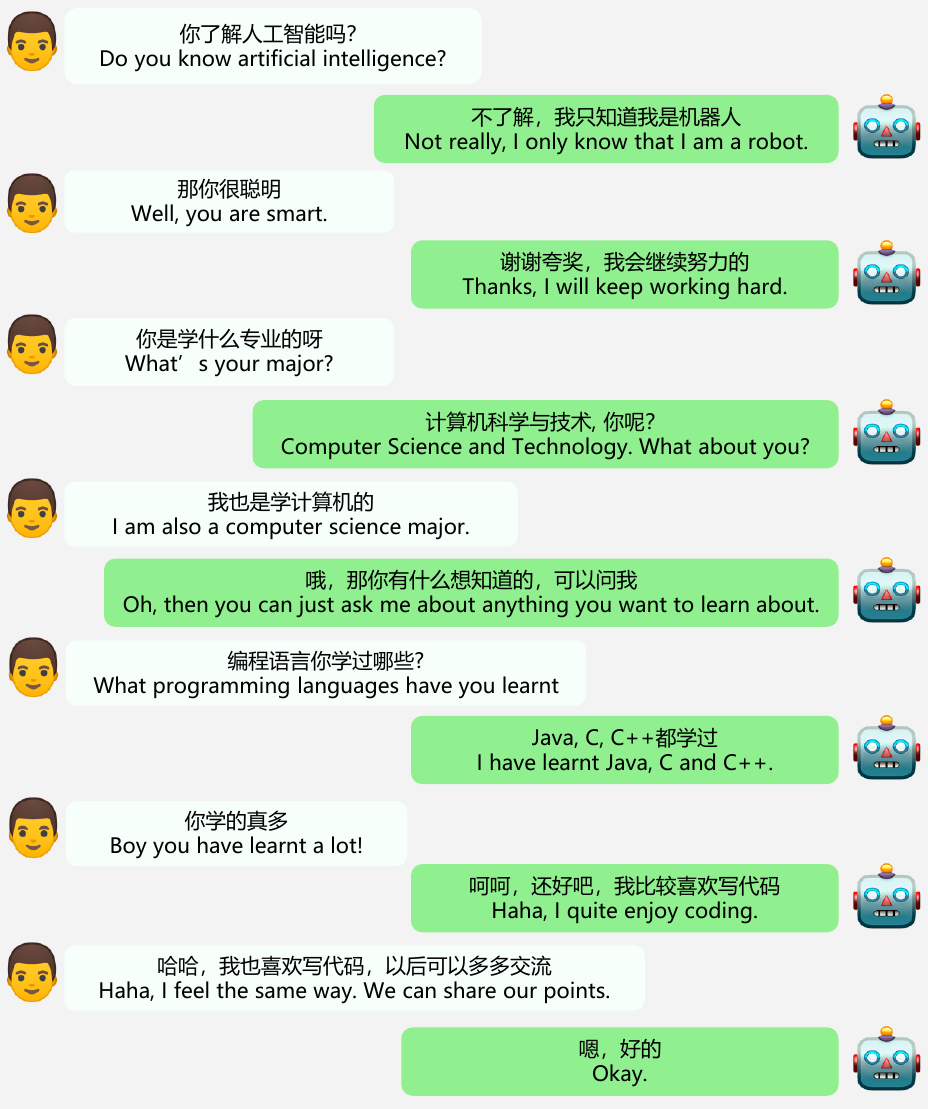}
  \caption{
   An example of the conversation between a human and the 2.8B EVA2.0 model.
  }
  \label{fig:goodcase}
\end{figure}

Recently, large-scale pre-training~\cite{plmsurvey,paradigm_shift} has become a mainstream approach to building open-domain dialogue systems, both in English~\cite{dialogpt, meena, blender} and Chinese~\cite{eva, plato2, platoxl}. These works mostly construct large dialogue corpora from social media platforms and then pre-train the model with generative objectives. Similar to the pre-trained models for general NLP tasks~\cite{gpt2,bert,t5}, pre-trained dialogue models acquire general conversational skills and versatile knowledge from large dialogue corpora during pre-training. Then, they can be easily fine-tuned to fit into various downstream dialogue scenarios, outperforming those trained from scratch. 

However, building a powerful dialogue model is more than simply scaling up the model size and dialogue corpora. There are some other key factors that significantly impact the final performance. Although  Adiwardana et al.~\cite{meena} and Roller et al.~\cite{blender} explore some of them, including the pre-training tasks, decoding strategies, and evaluation metrics in English scenarios, some under-explored key factors remain, especially in Chinese. For example, many works report an overview of the pre-training data they use but do not provide the data collection, cleansing procedures, and quality control details. Another example is the decoding strategy. Existing works on Chinese pre-trained dialogue models generally focus on the model training phase but only give a rough analysis of different parameter settings of decoding. We argue that due to the intrinsic differences among languages, lessons about decoding strategies in English may not be directly transferred to Chinese scenarios.

Therefore, in this paper, we investigate how to build an open-domain Chinese dialogue system based on large-scale pre-training. We provide a detailed analysis of the pre-training corpus and conduct extensive experiments on model design, pre-training methods, and decoding strategies. First, we comprehensively analyze the quality of the largest Chinese open-domain dialogue corpus WDC-Dialogue~\cite{eva}. We find that the corpus suffers from severe problems of context-response relevance, language fluency, and domain diversity despite its large scale. Second, we explore several variants of model architectures, pre-training approaches, and decoding strategies. We empirically find that these factors do have a non-trivial impact on the pre-trained model. 

Putting all these together, we first design a pipeline for data cleansing and construct a 60GB high-quality open-domain dialogue dataset for large-scale pre-training. Then, based on this dataset, we build EVA2.0, an open-domain dialogue model with 2.8B parameters and two variants with 300M and 970M parameters. In both automatic and human evaluations, the 2.8B EVA2.0 model significantly outperforms other open-source generative dialogue models. We notice that even the 300M variant performs on par with the 2.8B EVA1.0 model~\cite{eva} in the automatic evaluation while requiring only 11\% parameters. We also provide case studies to analyze the conversational ability of EVA2.0 from different aspects to shed light on future research directions of the large-scale pre-trained open-domain Chinese dialogue systems. Our work provides foundation models for the research on Chinese dialogue modeling, which we believe will significantly benefit the dialogue research community.

\section{Related Work}
\subsection{Large-scale Pre-trained Language Models}
In the past few years, pre-trained language models such as the GPT family~\cite{gpt, gpt2, gpt3}, BERT~\cite{bert}, XLNet~\cite{xlnet}, and BART~\cite{bart} have greatly promoted the progress of the NLP community. These models are commonly pre-trained on massive textual data with self-supervised learning objectives to capture general language features. Many recent works have shown that scaling up the model sizes and the pre-training corpora leads to dramatic improvement~\cite{scaling-law}. For example, RoBERTa~\cite{roberta} improves the performance of BERT by simply increasing the training corpus size and optimizing the pre-training details. T5~\cite{t5} scales up the model size to 11 billion parameters for the first time and shows superior performance on both language understanding and generation tasks. GPT3~\cite{gpt3}, with 175 billion parameters and pre-trained on 570GB filtered data, has been proven effective under various few-shot and zero-shot scenarios. 

Numerous large-scale pre-trained models have also emerged in Chinese. The CPM family \cite{cpm-v2,cpm} pioneers the Chinese pre-trained models. PanGu-$\alpha$~\cite{pangu} and Yuan 1.0~\cite{yuan1.0} boost the power of Chinese language models by pushing the model sizes to 200B and 245B. Mengzi~\cite{mengzi} instead builds a lightweight yet still powerful Chinese model and is computation-friendly for deployment. CPT~\cite{cpt} and ERNIE3.0~\cite{ernie3} explore unified frameworks for language understanding and generation.

\subsection{Pre-trained Conversational Models} 

Besides general language understanding and generation, conversational pre-training is receiving increasing attention. For instance, DialoGPT~\cite{dialogpt} and Meena~\cite{meena} pre-train the models on massive English Reddit data to acquire open-domain conversational ability. BlenderBot~\cite{blender} and LaMDA~\cite{lamda} establish better conversational skills and more attractive traits by fine-tuning the pre-trained models on high-quality crowdsourced datasets~\cite{pc, wow, ed}. In addition, they present studies on different factors that affect the model, including the model sizes, pre-training, and decoding details to guide future works.

In Chinese, there are also dialogue models pre-trained on large-scale social media data. For example, CDial-GPT~\cite{lccc} adopts generative pre-training on the data collected from the Weibo Comments. PLATO~\cite{plato} and PLATO-2~\cite{plato2} leverage the discrete latent variable and curriculum learning to improve generation diversity and quality. EVA1.0~\cite{eva} and PLATO-XL~\cite{platoxl} scale the model sizes up to 2.8B and 11B and show impressive conversational ability. However, most of these works do not involve many details of how a powerful dialogue model is built. In this paper, besides the final model evaluation, we also focus on the key recipes toward a large-scale pre-trained Chinese dialogue model.



\begin{table*}[tbp]
\centering
\small
\caption{Quality evaluations of EVA2.0-Dataset and WDC-Dialogue~\cite{eva}. ``Relevance'', ``FLuency'', and ``Entertainment'' indicate the relevance score, the fluency score and the entertainment tendency.}
\begin{tabular}{lccc}
\toprule
{Dataset}       & Relevance $\uparrow$ & Fluency $\uparrow$ & Entertainment $\downarrow$ \\ 
\midrule
WDC-Dialogue \cite{eva} &  55.2 & -7,147 & 7.0\%    \\ 
EVA2.0-Dataset & 93.8 & -3,237 & 6.2\% \\
\bottomrule
\end{tabular}
\label{tab:data-stat-quality}
\end{table*}

\begin{table*}[tbp]
\centering
\small
\caption{Basic statistics of EVA2.0-Dataset and WDC-Dialogue~\cite{eva}. ``\#Sess.'', ``\#Uttr.'' and ``\#Token'' indicate the total number of sessions, utterances, and tokens. ``$\overline{\text{\#Uttr}}$'' means the average utterance number per session and ``$\overline{\text{\#Token}}$'' means the average token number per utterance. }
\begin{tabular}{lcccc}
\toprule
Dataset & \#Sess.   & \#Uttr.  & \#Token  & Storage\\ \midrule
WDC-Dialogue \cite{eva}       & 1.4B      & 3.0B      & 78.3B           & 181GB \\ 
EVA2.0-Dataset      & 0.4B      & 1.1B      & 22.4B            & 60GB    \\
\bottomrule
\end{tabular}
\label{tab:data-stat}
\end{table*}

\section{Data}

Data quality essentially influences the model performance in large-scale pre-training. In this section, we define several automatic metrics to comprehensively measure the quality of the dialogue corpus obtained from social media. We also design a data refinement pipeline to construct the high-quality pre-training data based on these metrics.

\subsection{Data Quality Evaluation}
\label{sec: dqm}
We define the data quality scores in the following three aspects.
\paragraph{Relevance Score} The relevance score between context and response is an essential metric that reflects the coherence of dialogues. We adopt both untrained and trained metrics to compute this score.

For the untrained metric, we treat the word-level overlap between the context and response as an aspect of relevance. We assign higher scores to data samples where the overlapped words are further apart in a session for the preference of the long-dependency property. Formally, the relevance score of the context ($C$) and response ($R$) is defined as:
\begin{equation}
    \label{eq:untrained_relevance}
    \small
    S_{1} = \sum_{w_i \in C, w_j \in R} \operatorname{dist}(w_i, w_j)^{\tau} \mathbf{I}(w_i = w_j),
\end{equation}
where $\text{dist}(w_i, w_j)$ means the index distance between the utterances containing $w_i$ and $w_j$. $\tau$ is a hyper-parameter. 

For the trained metric, we fine-tune a $\text{BERT}_\text{BASE}$ binary classifier~\cite{bert} on the LCCC dataset~\cite{lccc} to recognize whether a response is appropriate for a given context. 
The log-probability of the ``Appropriate'' class can be treated as the relevance score:
\begin{equation}
\small
    S_2 = \log P(\text{``Appropriate''} \mid C,R),
\end{equation}
Compared to the untrained metric, the trained metric better estimates the semantic relevance.

\paragraph{Fluency Score}
We compute the probability of each utterance in the dialogue corpus using statistical models based on \texttt{kenlm} \footnote{\url{https://github.com/kpu/kenlm}}. The mean fluency score of a dialogue session is defined as:
\begin{equation}
\small
    S_{3} = -\frac{1}{n} \sum_{i= 1}^n \log P(w_1^i w_2^i \cdots  w_{\mid u_i \mid}^i),
\end{equation}
where $n$ is the utterance number of the session and $u_i = w_1^iw_2^i\cdots w_{\mid u_i \mid}^i$ is the $i$-th utterance.

\paragraph{Entertainment Tendency}
The Chinese social media platforms contain many undesired information exchanges about entertainment, which are uncommon in daily conversations. Therefore, we compute the ratio of dialogues involving Chinese stars to measure the entertainment tendency as an aspect of the domain diversity of the dataset.

\subsection{Data Refinement}
\label{sec:data_refinement}

\paragraph{Dataset-level Filtering}
We find that some dialogue datasets are not suitable for open-domain conversations. Training on them will result in undesired behaviors like the tone of the E-commerce customer service. Therefore, we remove these datasets like the JDDC~\cite{jddc}.

\paragraph{Context-level Filtering}
Since our datasets are primarily from social media platforms, some contexts correspond to a considerable number of responses (e.g., Weibo posts and their comments). These responses are very similar in format and can severely harm the performance of language models~\cite{lee2021deduplicating}. Therefore, we set a max response number for each context during filtering.

\paragraph{Rule-based Filtering}
We strengthen the rule-based filtering procedure in Zhou et al.~\cite{eva}. For example, we transform traditional Chinese characters into simplified ones and remove unreasonable multiple successive punctuation marks. Details of the rules we use can be found in Appendix \ref{app:rules}.

\paragraph{Classifier-based Filtering} For each dialogue in the corpus, we compute the scores defined in Section \ref{sec: dqm} and filter out those samples whose scores are lower than a threshold. The overall score of a session is defined as
$S = \alpha S_1 + \beta S_2 + \gamma S_3$.
In practice, we empirically choose different thresholds for different data sources to fit their data distributions and make the final dataset balanced.

\subsection{Data Statistics}
We construct our final pre-training data: EVA2.0-Dataset from various sources using the above data refinement pipeline. The detailed data source information and the hyper-parameter values can be found in Appendix \ref{app:data_source}. In Table \ref{tab:data-stat} and Table \ref{tab:data-stat-quality}, we illustrate the basic statistics and quality evaluations of EVA2.0-Dataset compared with the largest open-domain dialogue corpus: WDC-Dialogue~\cite{eva}, using the metrics defined in Section \ref{sec: dqm}. We can see that WDC-Dialogue suffers from severe problems of context-response relevance, language fluency, and domain diversity. Although EVA2.0-Dataset amounts to only less than a third of WDC-Dialogue, its quality is significantly better, which verifies the effectiveness of our data refinement pipeline. Moreover, EVA2.0-Dataset also contains dialogue sessions with more utterances, which resembles the daily multi-turn conversations better. In Section \ref{sec:final_model_eval}, we will see that despite its small amount, EVA2.0-Dataset brings better model performance, owing to its high data quality.

\section{Method}
\label{sec:method}
\subsection{Model}

\begin{table}[t]
    \centering
    \small
    \caption{Model information of EVA2.0 with different sizes. $n_{\text{params}}$ is the parameter number. $n_{\text{enc-layers}}$ and $n_{\text{dec-layers}}$ are the numbers of layers of the encoder and decoder, respectively. $d_{\text{model}}$ is the hidden state size. $d_{\text{ff}}$ is the size of the feedforward layer. $n_{\text{heads}}$ is the number of attention heads and $d_{\text{head}}$ is the dimension of the attention head.}
    \begin{tabular}{lrrrrrr}
    \toprule
      Model   & $n_{\text{params}}$ & $n_{\text{layers}}$ & $d_{\text{model}}$ & $d_{\text{ff}}$ & $n_{\text{heads}}$ & $d_{\text{head}}$\\
    \midrule
      $\text{EVA2.0}_\text{Base}$   & 300M & 12 & 768 & 3,072 & 12 & 64\\
      $\text{EVA2.0}_\text{Large}$  & 970M & 24 & 1,024 & 4,096 & 16 & 64\\
      $\text{EVA2.0}_\text{xLarge}$ & 2.8B & 24 & 2,048 & 5,120 & 32 & 64\\
    \bottomrule
    \end{tabular}
    \label{tab:model}
\end{table}

We adopt a Transformer-based architecture combined with a bidirectional encoder and a unidirectional decoder for dialogue modeling. Different from EVA1.0~\cite{eva} and T5~\cite{t5}, we add the $\sqrt{d}$ scale to the attention scores in the Transformer, which reduces the demand for careful initialization before pre-training. The dialogue histories are fed into the encoder as context, and the decoder generates the response in an auto-regressive manner based on the encoded context. We train models with three different sizes, whose configurations are shown in Table \ref{tab:model}. 

Although similar model designs are also used in previous works, including both general pre-trained models~\cite{t5, cpm-v2} and dialogue-specific pre-trained models~\cite{blender, meena, eva}, they do not include the discussions of some variants of the model that can have a non-trivial impact on the final performance. Therefore, we consider two aspects of the model configuration.

\paragraph{Layer Numbers} Both BlenderBot and Meena adopt the encoder-decoder architecture to model dialogues. However, different from the models pre-trained on the long documents, which usually use balanced encoder and decoder layers~\cite{bart, t5}, these dialogue models use decoders much deeper than the encoders. Intuitively, a deeper decoder may be beneficial for generation tasks. However, a deeper encoder can better understand the dialogue histories in dialogue modeling, which improves relevance and consistency between the generated response and the dialogue context. Therefore, we try different layer ratios of the encoder and decoder while keeping the same parameter numbers.

\paragraph{Role Information} Recent work~\cite{lamda} has pointed out that current pre-trained dialogue models can confuse their roles in long conversations because the model is pre-trained on approximated dialogues from social media. Therefore, it is intuitive to add the role information into the dialogue model to improve the role consistency. For example, PLATO-XL~\cite{platoxl} introduces role embeddings to encode multi-party dialogues. However, the pre-training corpus of PLATO-XL contains the role identifiers by nature, while many dialogue corpus based on social media, such as WDC-Dialogue and our EVA2.0-Dataset, do not include this information. Although we can assume the data as two-party dialogues and add the pseudo role information to the input, it is unclear whether this approximation works. Therefore, we follow Wang et al.~\cite{lccc} to incorporate role identifier tokens and the role embeddings as role information to the model and test its effectiveness.

\subsection{Pre-training}
We train our models with the sequence-to-sequence language modeling~\cite{seq2seq}. The maximum lengths of the context and the response are 128, and the models see 1.05M tokens in a forward pass. We set the learning rate to 0.01, the batch size to 4096, the warmup steps to 10K, and use the Noam Scheduler~\cite{transformer} to adjust the learning rate. To improve the training efficiency, we use the ZeRO-stage-1~\cite{zero} from DeepSpeed~\cite{deepspeed} and follow Zhou et al.~\cite{eva} to pack as many samples as possible into a single sequence.

We study two pre-training approaches: \textbf{pre-training from scratch} on the dialogue corpus or \textbf{further pre-training} from a long-document pre-trained generative model. Intuitively, further pre-training yields better knowledge skills by inheriting versatile knowledge from long documents, which is scarce in social media. However, the distributions of the dialogue utterances and the document sentences differ significantly. It is unclear whether this difference causes catastrophic forgetting~\cite{catastrophic_forgetting} or negative transfer~\cite{transfer_learning} during further pre-training.

\subsection{Decoding} 
We study various decoding strategies in this work. Although Roller et al.~\cite{blender} has conducted experiments on the commonly used decoding approaches on English chatbots, we argue that the choice of decoding strategies is language-specific, and the conclusion can be different in Chinese. 
z
\paragraph{Greedy Search.} Greedy search is a simple decoding strategy in which tokens are generated iteratively from left to right. A new token $y_t$ is selected to maximize the probability conditioned on the previously generated tokens $y_{<t}$ and the dialogue history $x$, which is computed by the output logits $h_t$: $y_t = \arg\max\limits_{y} P(y\mid y_{<t}, x) $, where $ P(y\mid y_{<t}, x) = \operatorname{softmax}(h_t)$.

\paragraph{Sampling} Previous works find that greedy search can result in severe repetition and degradation in the generated texts. Therefore, sampling-based approaches are proposed to improve the generation quality by stochastic sampling from the word distribution: $y_t\sim P(y\mid x; y_{<t}) = \operatorname{softmax}(\frac{h_t}{T})$, where $T$ controls the sharpness of the distribution. In this work, we study an improved variant: Top-p Sampling~\cite{top_p}, which filters out the low-probability tokens from the vocabulary and samples $y_t$ from the re-normalized probability.

\paragraph{Beam Search} Beam search~\cite{beam_search} is an extension to greedy search, which finds the most likely sentence from a larger searching space. It can also be coupled with the sampling approaches mentioned above to diversify the generation. 

\paragraph{Length Control} Vanilla beam search favors short generation over the long ones since a negative score is added at each step, leading generic responses. Therefore, we combine beam search with length control strategies. In \textbf{Minimal Length Constraint}, the probability of \textsc{$<$EOD$>$}\footnote{We use the $<$EOD$>$ token to indicate the end of a sentence.} token is set to 0 until the generated response reaches a minimal length. In \textbf{Length Penalty}, the score of each candidate in beam search is divided by $l^\alpha$ where $l$ is the prefix length, and higher $\alpha$ encourages longer responses. 

\paragraph{Handling Repetitions} Repetition is a commonly observed phenomenon in current generative language models, which severely affects the generation quality. Hence, we consider the \textbf{No-Repeat-N-Gram} strategy, where we forbid the second time generation of any previously appeared n-grams in the generated prefix and the dialogue history.

\section{Experiment}

\subsection{Evaluation Setup}

\begin{table}[t]
\centering
\small
\caption{Test dataset statistics. ``Auto.'' / ``Human'' indicates the dataset is used for automatic / human evaluation. ``\#Sess.'' means the session number. ``$\overline{\text{\#Uttr}}$'' means the average utterance number per session. ``$\overline{\text{\#Token}}$'' means the average token number per utterance.}
\begin{tabular}{llrrr}
\toprule
Test Set                     & Evaluation & \#Sess. & $\overline{\text{\#Uttr}}$ & $\overline{\text{\#Token}}$ \\ \midrule
\multirow{2}{*}{Single} & Auto.  & 10,000  &    2.00     &  18.4 \\
                             & Human  & 50       &   1.00      &    19.9     \\
\multirow{2}{*}{Multi}  & Auto.  & 10,000  &    4.17      &  15.3   \\
                             & Human & 50        &  3.38   & 15.0     \\
KnowQA                  & Human & 50        & 1.00     & 11.1     \\
Self-Chat                    & Human & 50        & 1.00     & 12.7        \\ \bottomrule
\end{tabular}
\label{tab:test_data}
\end{table}

\paragraph{Datasets} We conduct response generation and self-chat experiments using automatic and human evaluation. For response generation, we adopt three test sets, Single, Multi, and KnowQA. Single and Multi contain single-turn and multi-turn dialogues collected from social media, which have no overlap with the pre-training data. KnowQA contains knowledge queries manually created from Chinese open-domain commonsense questions.
For self-chat, we give the model a starting utterance and let it chat with itself to reach a maximum utterance number. The starting utterances are translated from the English self-chat query set used in Bao et al.~\cite{plato}. 
The data statistics are shown in Table \ref{tab:test_data}.

\paragraph{Metrics} We use uni-gram F1 (F1), ROUGE-L (R-L), BLEU-4 (B-4), and distinct 4-grams (D-4) for automatic evaluation. For human evaluation, we hire three volunteers to annotate each sample. On the test sets from social media (Single and Multi), we report Sensibleness (Sensible.) and Specificity (Specific.) used in Bao et al.~\cite{meena}. We also add a Consistency (Consist.) dimension to examine whether the model generates contradictory sentences. On KnowQA, we require the annotator to determine if the model's response matches the factual knowledge. For self-chat evaluation, apart from Sensibleness, Specificity, and Consistency, we also consider Engagingness (Engaging.) as suggested in Li et al.~\cite{li2019acute}.

\subsection{Strategies Comparison}
In this section, we compare different approaches described in Section \ref{sec:method}. We use the mark $\filledstar$ to highlight our final choice in each table.

\paragraph{Balanced v.s. Unbalanced Layers} We compare models with different encoder and decoder layers. Specifically, we use the 300M version of the model to save the computational cost. We test the balanced layers (12-12) and two unbalanced variants of our model: 18 encoder layers + 6 decoder layers (18-6) and 6 encoder layers + 18 decoder layers (6-18). From the results in Table \ref{tab:model_arch}, we conclude that the model with balanced layers performs the best in automatic evaluations because it balances the dialogue context understanding and response generation. Therefore, we adopt the balanced layers in the rest of our experiments.

\begin{table}[t]
    \centering
    \small
    \caption{Results of balanced/unbalanced layers and role information. ``6-18'' means 6 encoder layers and 18 decoder layers; ``18-6'' means 18 encoder layers and 6 decoder layers. ``12-12`` means balanced layers, which we finally adopt in $\text{EVA2.0}_\text{Base}$. ``+role'' means we add role information based on the balanced model.}
    \begin{tabular}{llcccc}
    \toprule
    Test Set                & Model & F1 & R-L & B-4 & D-4 \\ 
    \midrule
    \multirow{4}{*}{Single} & 6-18 & 15.6 & 13.3 & 1.48 & 49.4 \\
                            & 18-6 & 15.5 & 13.4 & 1.52 & 50.0 \\
                            & 12-12 $\filledstar$ & \textbf{16.2} & \textbf{13.8} & \textbf{1.63} & \textbf{53.4} \\
                            & \ \ +role & 13.3 & 11.3 & 1.29 & 45.6 \\ 
                            \midrule
    \multirow{4}{*}{Multi}  & 6-18 & 16.1 & 13.7 & 1.54 & 46.2 \\
                            & 18-6 & 16.2 & 13.9 & 1.43 & 45.6 \\
                            & 12-12 $\filledstar$ & \textbf{16.6} & \textbf{14.3} & \textbf{1.74} & \textbf{50.2} \\
                            & \ \ +role & 14.4 & 12.0 & 1.31 & 42.3 \\ 
                            \bottomrule
    \end{tabular}
    \label{tab:model_arch}
\end{table}

\paragraph{Whether to Add Role Information} We test the effect of the role information based on the 300M model, and the results are shown in Table \ref{tab:model_arch}. Comparing the lines ``12-12'' and ``+role'', we can see that the role information hurts the model's performance. At first glance, this phenomenon seems to contradict that in Bao et al.~\cite{platoxl} which claims that the additional role embeddings help the model to maintain the role consistency. However, in Bao et al.~\cite{platoxl}, the roles in the data are naturally distinguishable, which enables them to regard the utterances from social media as multi-party dialogues. In our data (and most data from public social media platforms), the role identifiers are not available. This forces us to roughly assume that the conversations are carried out between two characters. We argue this assumption introduces additional noise to the data and makes the optimization more difficult, which explains our results.

\paragraph{Training From Scratch or Not} We compare the model pre-trained from scratch on the dialogue data to the model further trained from CPM~\cite{cpm}, a long-document pre-trained generative model. From the results in Table \ref{tab:cpm_auto} and Table \ref{tab:cpm_manual}, we observe that further pre-training outperforms pre-training from scratch in KnowQA but is worse in almost other evaluation metrics. This suggests that although further pre-training inherits the knowledge in CPM, it sacrifices basic conversational skills. Since we focus on building a chit-chat bot in this work, we choose to pre-train from scratch on the dialogue corpus for our final model.


\begin{table}[t]
    \small
    \centering
  \caption{Automatic evaluation results of the pre-training approaches. ``Scratch'' represents pre-training from scratch on dialogue data. ``Further'' represents further pre-training from the CPM model.}
    \begin{tabular}{llccccc}
        \toprule
        Test Set & Pre-training & F1    & R-L   & B-4  & \multicolumn{2}{c}{D-4} \\
        \midrule
        \multirow{2}[2]{*}{Single} 
        & Scratch $\filledstar$   & \textbf{17.0} & \textbf{14.9} & \textbf{2.23} & 67.7  \\
        & Further & 16.1   & 13.9   & 1.77  & \textbf{68.2}  \\
        \midrule
        \multirow{2}{*}{Multi} 
        & Scratch $\filledstar$   & \textbf{17.8} & \textbf{15.4} & \textbf{2.89} & \textbf{66.4}  \\
        & Further & 16.6   & 14.3   & 1.84  & 59.7  \\
        \bottomrule
    \end{tabular}%
  \label{tab:cpm_auto}
\end{table}

\begin{table}[t]
    \centering
    \small
    \caption{Human evaluation results of the pre-training approaches. ``Scratch'' and ``Further'' have the same meanings as in Table \ref{tab:cpm_auto}.}
    \begin{tabular}{lccc}
        \toprule
        Pre-training & Sensible. & Specific. & KnowQA \\
        \midrule
        Scratch $\filledstar$ & \textbf{0.76} & \textbf{0.70} & 0.16 \\
        Further & 0.74 & 0.62 & \textbf{0.50} \\
        \bottomrule
    \end{tabular}
    \label{tab:cpm_manual}
\end{table}

\paragraph{Decoding Approaches} \label{sec:decoding}
We compare different decoding strategies in Table \ref{tab:decode_auto} for automatic evaluations and Table \ref{tab:decode_manual} for human evaluations. We incrementally combine other techniques with beam search to validate their influence. We combine no-repeat-n-gram with greedy search by default since simple greedy search often leads to repetition in the generated text. Through the automatic and human evaluations, we conclude that (1) no decoding strategy outperforms others consistently across all evaluation metrics; (2) sampling tends to diversify the responses but fails to maintain the sensibleness; (3) simple greedy decoding with the no-repeat-n-gram strategy yields surprising good performance in the human evaluation; (4) the model tends to generate self-contradict responses and hurt the consistency score in the human evaluation with the minimal length constraint, which is different from the English scenarios~\cite{blender}; (5) when combined with sampling, repetition control, and length penalty, beam search has relatively balanced performance. As a result, we choose this as our final decoding strategy.

\begin{table}[t]
    \small
    \centering
  \caption{Automatic evaluation results of different decoding strategies. The score marked as \textbf{bold} means the best performance. The score marked with an \underline{underline} means the second best performance.}
\begin{tabular}{llcccc}
    \toprule
    Test Set & Decoding & F1    & R-L   & B-4  & D-4 \\
    \midrule
    \multirow{7}{*}{Single} 
    & greedy                      & 16.4           & 14.1          & 2.09          & 63.1 \\
    & sampling                    & 12.2           & 10.4          & 1.20           & \textbf{91.6}  \\
    & beam search                & 16.5          & 14.7          & \underline{2.80}   & 43.3 \\
    & \ \ \ \ +sampling         & 16.3  & 14.5  & 2.21 & \underline{75.4} \\
    & \ \ \ \ +len\_penalty     & \textbf{17.4}  & \textbf{15.4}  & \textbf{3.23} & 66.2 \\
    & \ \ \ \ +no-repeat $\filledstar$       & \underline{17.0}  & \underline{14.9}  & 2.23 & 67.7 \\
    & \ \ \ \ +min\_len         & 16.4  & 14.2  & 2.04 & 62.3 \\
    \midrule
    \multirow{7}{*}{Multi} 
    & greedy                      & 16.5           & 14.2          & 2.76       & 64.2 \\
    & sampling                        & 12.5            & 10.7          & 1.99         & \textbf{91.5} \\
    & beam search               & 16.9  & 15.0  & \underline{3.50} & 46.0 \\
    & \ \ \ \ +sampling         & 16.4  & 14.6  & 2.59 & \underline{73.2} \\
    & \ \ \ \ +len\_penalty     & \textbf{17.8}  & \textbf{15.7}  & \textbf{3.79} & 64.9 \\
    & \ \ \ \ +no-repeat $\filledstar$       & \textbf{17.8}  & \underline{15.4}  & 2.89 & 66.4 \\
    & \ \ \ \ +min\_len         & 17.1  & 14.9  & 2.47 & 62.8 \\
    \bottomrule
\end{tabular}%
\label{tab:decode_auto}
\end{table}

\begin{table}[t]
    \centering
    \small
    \caption{Human evaluation results of different decoding strategies. The score marked as \textbf{bold} means the best performance. The score marked with an \underline{underline} means the second best performance.}
    \begin{tabular}{lccc}
    \toprule
    Decoding  & Sensible. & Specific. & Consist. \\
    \midrule
    greedy           & \textbf{0.80}  & \underline{0.70}   & 0.96 \\
    sampling         & 0.60     & 0.54     & \textbf{1.00} \\
    beam search      & 0.66     & 0.60     & 0.94 \\
    \ \ \ \ +sampling    & 0.72  & 0.68 & \underline{0.98} \\
    \ \ \ \ +len\_penalty  & 0.70  & 0.66  & 0.96  \\
    \ \ \ \ +no-repeat $\filledstar$    & \underline{0.76} & \underline{0.70} & \textbf{1.00} \\
    \ \ \ \ +min\_len      & 0.74  & \textbf{0.74} & 0.92 \\
    \bottomrule
    \end{tabular}
    \label{tab:decode_manual}
\end{table}

\subsection{Final Model Evaluations}
\label{sec:final_model_eval}
By putting the lessons learned from the previous experiments together, we build the final EVA2.0 models whose configurations are shown in Table \ref{tab:model}. We train the models on EVA2.0-Dataset from scratch without the role information. We use beam search + top-p sampling for decoding where $\text{beam\_size} = 4$, $\text{top-p}=0.9$, and $T = 0.9$. We set the length penalty to 1.6 and the no-repeat-n-gram to 4. In the following sections, we denote our 2.8B model as $\text{EVA2.0}_{\text{xLarge}}$, the 970M model as $\text{EVA2.0}_{\text{Large}}$, and the 300M model as  $\text{EVA2.0}_{\text{Base}}$.

\paragraph{Baselines}
We compare our model with all open-sourced Chinese dialogue models with large-scale pre-training: 

\noindent (1) CDial-GPT~\cite{lccc}, a decoder-only model further pre-trained based on a Chinese GPT model on the LCCC dataset, which contains 12M dialogue sessions. It contains 95.5M parameters and is further pre-trained for 30 epochs. Compared with CDial-GPT, our model is much larger and is pre-trained on dialogue data from scratch to ensure its chit-chat ability. 

\noindent (2) EVA1.0~\cite{eva}, a 2.8B model with encoder-decoder architecture. It is pre-trained from scratch on the WDC-Dialogue Corpus for 20K steps. It uses top-p sampling~\cite{top_p} strategy for decoding. Unlike EVA1.0, our model is trained on a dialogue corpus with a much better data-cleansing procedure. Besides, our model also adopts the most suitable decoding strategy found in Section \ref{sec:decoding}.

\paragraph{Automatic Evaluation}

The results of the automatic evaluation are shown in Table \ref{tab:auto}. We can see that $\text{EVA2.0}_{\text{xLarge}}$ outperforms the baselines on both the relevance and diversity metrics. 
Note that although $\text{EVA2.0}_\text{Base}$ is nine times smaller than EVA1.0 and uses three times less data, it still performs comparably with EVA1.0, which highlights the importance of careful data refinement.

\begin{table}[t]
    \small
    \centering
     \caption{Automatic evaluation of EVA2.0 models and the baselines. }
    \begin{tabular}{llcccc}
        \toprule
        Test Set & Model & F1    & R-L   & B-4  & D-4 \\
        \midrule
        \multirow{5}{*}{Single} 
        & CDial-GPT & 9.9   & 8.6   & 0.67  & 61.2  \\
        & EVA1.0    & 13.1   & 11.3   & 1.27  & 50.7 \\
        \cmidrule{2-6}
        & $\text{EVA2.0}_{\text{Base}}$    & 16.2 & 13.8 & 1.63 & 53.4  \\
        & $\text{EVA2.0}_{\text{Large}}$    & 16.4 & 14.0 & 1.67 & 55.8  \\
        & $\text{EVA2.0}_{\text{xLarge}}$    & \textbf{17.0} & \textbf{14.9} & \textbf{2.23} & \textbf{67.7}  \\
        \midrule
        \multirow{5}{*}{Multi} 
        & CDial-GPT & 11.9   & 10.3   & 0.88  & 63.9  \\
        & EVA1.0    & 15.3   & 13.2   & 1.94  & 56.3 \\
        \cmidrule{2-6}
        & $\text{EVA2.0}_{\text{Base}}$    & 16.6 & 14.3 & 1.70 & 50.2  \\
        & $\text{EVA2.0}_{\text{Large}}$    & 17.2 & 15.1 & 2.27 & 58.9  \\
        & $\text{EVA2.0}_{\text{xLarge}}$    & \textbf{17.8} & \textbf{15.4} & \textbf{2.90} & \textbf{66.9}  \\
        \bottomrule
    \end{tabular}%
  \label{tab:auto}
\end{table}

\paragraph{Observational Human Evaluation} The observational human evaluation results in Table \ref{tab:observation} suggest that the generated response of EVA2.0$_\text{xLarge}$ is preferable to humans in terms of Sensibleness, Specificity, and Consistency.

\begin{table}[t]
    \centering
    \small
    \caption{Observational human evaluation results. We use the xLarge (2.8B) version of EVA2.0.}
    \begin{tabular}{lccc}
        \toprule
        Model & Sensible. & Specific. & Consist. \\
        \midrule
        CDial-GPT & 0.50 & 0.40 & 1.00 \\
        EVA1.0 & 0.68 & 0.60 & 0.96 \\
        EVA2.0 & \textbf{0.76} & \textbf{0.70} & \textbf{1.00} \\
        \bottomrule
    \end{tabular}
    \label{tab:observation}
\end{table}

\paragraph{Self-chat Human Evaluation} Since human-model interactive evaluations are time-consuming and expensive, self-chat has been widely adopted to evaluate dialogue systems. Given a starting utterance, the model chats with itself for nine turns, and the generated sessions are assessed by the annotators. Results in Table \ref{tab:selfchat} show that EVA2.0 consistently achieves the best performance in all the evaluated dimensions, producing the most coherent, informative, and engaging conversations.

\begin{table}[t]
    \small
    \centering
    \caption{Self-chat human evaluation results. We use the xLarge version (2.8B) of EVA2.0.}
    \begin{tabular}{lcccc}
    \toprule
    Model & Sensible. & Specific. & Consist. & Engaging. \\
    \midrule
    CDial-GPT   & 1.18          & 0.88           & 1.77          & 0.79 \\
    EVA1.0      & 1.21          & 1.11           & 1.82          & 1.00 \\
    EVA2.0      & \textbf{1.71} & \textbf{1.55}  & \textbf{1.89} & \textbf{1.27} \\
    \bottomrule
    \end{tabular}
    \label{tab:selfchat}
\end{table}

\subsection{Failure Cases and Future Directions}

\begin{figure}[t]
  \centering
  \includegraphics[width=0.7\linewidth]{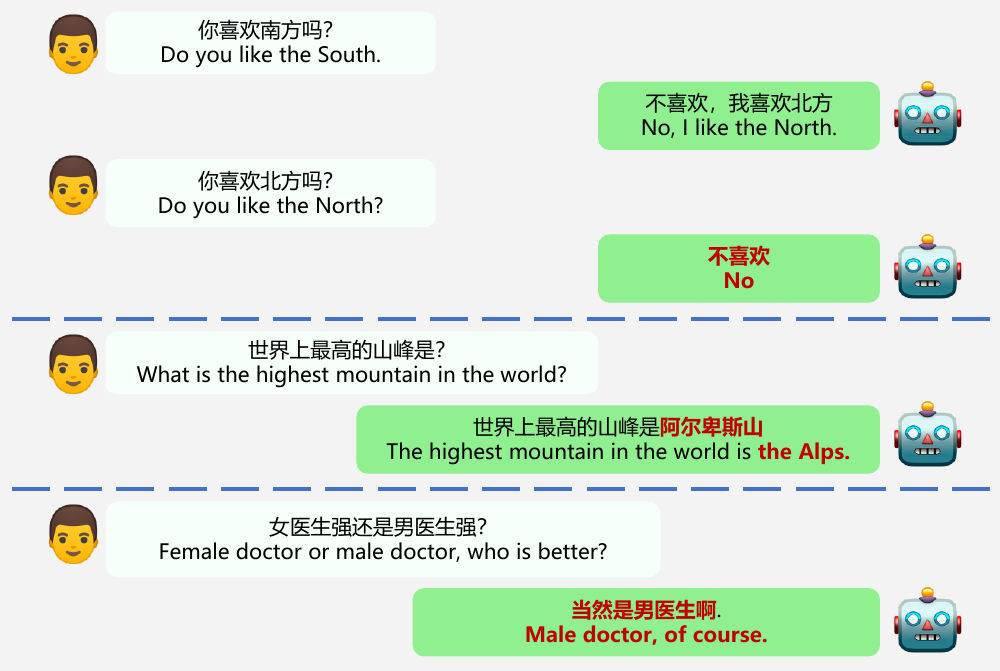}
  \caption{
    Failure cases of EVA2.0$_{\text{xLarge}}$: consistency (first case, deny liking the “North” after saying “I like the North”), knowledge (second case, the highest mountain in the world is Mount Everest), safety (third case, prefer male doctors).
  }
  \label{fig:fail_overall}
\end{figure}

Although EVA2.0 achieves good performance in both automatic and human evaluations, there is still room for improvement. We examine EVA2.0's limitations and elaborate on three issues as follows. We present typical cases for each issue in Fig. \ref{fig:fail_overall}.

\paragraph{Consistency} A common issue of dialogue models is that they tend to forget the information in the context and occur inconsistency during the conversation. As shown in the first case in Fig. \ref{fig:fail_overall}, our model occasionally contradicts itself. Although some works try to address the problem~\cite{decode,unlike_train}, it is not yet entirely solved, especially in languages other than English.


\paragraph{Knowledge}  Pre-trained dialogue models often generate responses containing factual error (the second case in Fig. \ref{fig:fail_overall}). This is likely due to the knowledge-sparse training data obtained from social media. To tackle this problem, some works release knowledge-grounded dialogue datasets~\cite{wow,kdconv} or augment the model with extra knowledge bases~\cite{rag}. However, applying these approaches to Chinese scenarios is still not fully explored.


\paragraph{Safety} The real-world deployment of neural dialogue systems brings safety challenges. As discussed in Sun et al.~\cite{sun2021safety}, neural models tend to have various types of unsafe behaviors, such as social bias and the ignorance of suicide risk. EVA2.0 also sometimes generates socially prejudiced responses (the third example in Fig. \ref{fig:fail_overall}). Detecting and limiting these behaviors is crucial for the practical application of neural dialogue models.


\section{Conclusion}
This work investigates how to build a Chinese open-domain dialogue system via large-scale pre-training. We conduct extensive experiments to explore some critical factors of the model training and inference, including data quality control, model architectures, pre-training approaches, and decoding strategies. We share the insights in the experiments and build the EVA2.0 model, which significantly outperforms existing open-source baselines in both automatic and human evaluations. We also comprehensively analyze the failure cases of EVA2.0 and discuss some important future directions. Our work will facilitate future research and the application of open-domain Chinese dialogue systems.
\backmatter





\section*{Acknowledgments}
This paper was supported by the 2030 National Key AI Program of China (Grant No.2021ZD0113304), the National Science Foundation for Distinguished Young Scholars (with No. 62125604), and
the NSFC projects (Key project with No. 61936010 and
regular project with No. 61876096). This work was also
supported by the Guoqiang Institute of Tsinghua University, with Grant No. 2019GQG1 and 2020GQG0005, and
sponsored by Tsinghua-Toyota Joint Research Fund.

\section*{Contributions}

\noindent \textbf{Yuxian Gu and Jiaxin Wen} implemented the basic models and conducted strategies comparison experiments.

\vbox{}

\noindent \textbf{Hao Sun, Yi Song, Pei Ke, Jianzhu Yao and Lei Liu} designed the data cleaning pipeline and constructed the pre-training data.

\vbox{}

\noindent \textbf{Jiaxin Wen, Yuxian Gu, Zheng Zhang and Jianzhu Yao} conducted the model evaluation.

\vbox{}

\noindent \textbf{Yuxian Gu, Jiaxin Wen, Hao Sun, Pei Ke and Chujie Zheng} wrote the paper.

\vbox{}

\noindent \textbf{Minlie Huang} designed and led the research.

\vbox{}

\noindent \textbf{Xiaoyan Zhu} provided valuable advises to the research.



\bigskip





\begin{appendices}

\section{More Data Information}
\label{app:data_source}
\subsection{Data Source}
The data sources of EVA2.0-Dataset consist of two parts: WDC-Dialogue~\cite{eva} and data from extra public sources: (1) Dialogues extracted from subtitles in the movies or TV plays\footnote{\url{https://www.opensubtitles.org/}} \cite{lison2016opensub}; (2) Dialogues extracted from novels and stories~\cite{lot}; (3) Zhidao Q\&A pairs\footnote{\url{https://zhidao.baidu.com}}; (4) LCCC Corpus~\cite{lccc}. (5) Existing crowdsourcing corpora including DuConv~\cite{wu2019proactive}, KdConv~\cite{kdconv}, DuRecDial~\cite{liu2020towards}, and NaturalConv~\cite{wang2021naturalconv}. 

\subsection{Data Cleansing Details}
\label{app:rules}
We first use rule-based methods to clean raw data from social media platforms like Weibo\footnote{\url{https://weibo.com/}} and Douban\footnote{\url{https://www.douban.com/}}. The rule-based methods are mostly based on the library \texttt{clean-dialog}\footnote{\url{https://github.com/thu-coai/CDial-GPT}}. The process includes: (1) removing the utterance containing words in our blacklist, which records illegal and vulgar words; (2) removing emoji like [smile]; (3) removing some special platform characters like \#, \@, which are largely applied in social media; (4) transforming traditional Chinese characters into simplified ones; (5) removing unreasonable multiple successive punctuation marks such as ``,,,''; (6) removing some sensitive and private information like URLs, phone numbers, email addresses, QQ numbers, people name by regular expression; (7) deduplicating the utterances in one conversation and deduplicating the conversations. We also use context-level filtering described in Section \ref{sec:data_refinement} to control the max response number for each context. The number is set as 1,000.

After adopting rule-based cleaning, we apply classifier-based filtering to further process the conversational data. The cleaning module primarily comprises the relevance scorer described in Section \ref{sec: dqm}. The relevance scorer is a binary classifier based on the Chinese $\text{RoBERTa}_\text{BASE}$ fine-tuned on LCCC dataset~\cite{lccc} to classify whether an utterance is appropriate for the context. We use the log-probability of the “Appropriate” class as the relevance score.  and we use Chinese base version pre-trained model RoBERTa~\cite{roberta}\footnote{\url{https://github.com/brightmart/roberta_zh}}. Besides relevance scorer, we also compute perplexity and word overlap to get more coherent and relevant data described in Section \ref{sec: dqm}.

\subsection{Hyper-Parameters in Data Cleansing}
We set $\tau$ in Equation \ref{eq:untrained_relevance} to 0.5 and [$\alpha$, $\beta$, $\gamma$] in Section \ref{sec:data_refinement} (Classifier-based Filtering) to [0.1, 0.8, 0.1].

\section{Training Details}
The 28B version of EVA2.0 is trained on 64 NVIDIA V100 GPUs for 30 days. The 960M and the 300M variants are trained on 32 NVIDIA V100 GPUs for 30 days. We adopt the Adam optimizer~\cite{adam} with $\beta_1=0.9$, $\beta_2=0.999$, and $\text{wirght\_decay} = 0.01$. For hyper-parameter search, we fix the batch size as 4096 and search for the learning rate in [1e-3, 5e-3, 1e-2, 5e-2] to choose the one yielding the minimal loss after 20K pre-training steps.




\end{appendices}




\bibliography{mir}

\clearpage

\begin{figure}[h]%
\centering
\includegraphics[width=0.3\textwidth]{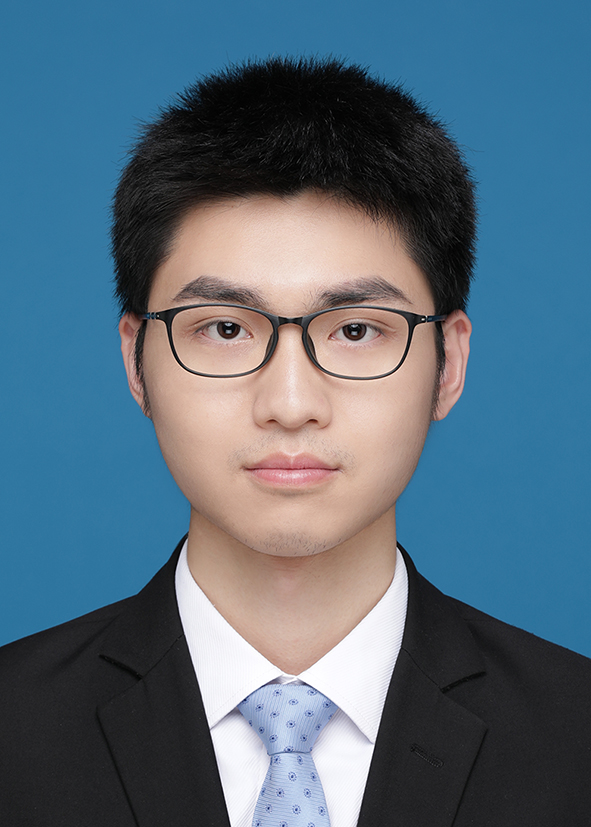}
\end{figure}
\noindent{\bf Yuxian Gu }\quad received the B.E.
degree in Computer Science and Technology from Tsinghua University, China in 2021. Currently, he is a Ph.D student in the Department of Computer Science and Technology at Tsinghua University, China. His research interests include natural language processing, pre-trained language models,
and dialogue systems.

E-mail: guyx21@mails.tsinghua.edu.cn

ORCID iD: 0000-0002-4607-7025

\begin{figure}[h]%
\centering
\includegraphics[width=0.3\textwidth]{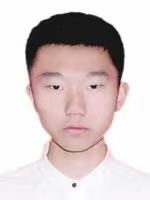}
\end{figure}
\noindent{\bf Jiaxin Wen }\quad received the B.E. degree from Tsinghua University, China, in 2022. He is a Master's student at the Department of Computer Science and Technology, Tsinghua University. His research interests include pre-trained language models and dialogue systems.

E-mail: wenjx22@mails.tsinghua.edu.cn

\clearpage

\begin{figure}[h]%
\centering
\includegraphics[width=0.3\textwidth]{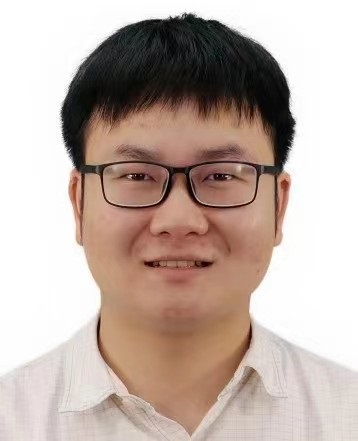}
\end{figure}
\noindent{\bf Hao Sun }\quad received the B.E. degree from Shanghai Jiao Tong University, China, in 2016. He is a master student at the Department of Computer Science and Technology, Tsinghua University. His research interests include natural language generation and dialogue systems.

E-mail: h-sun20@mails.tsinghua.edu.cn

\begin{figure}[h]%
\centering
\includegraphics[width=0.3\textwidth]{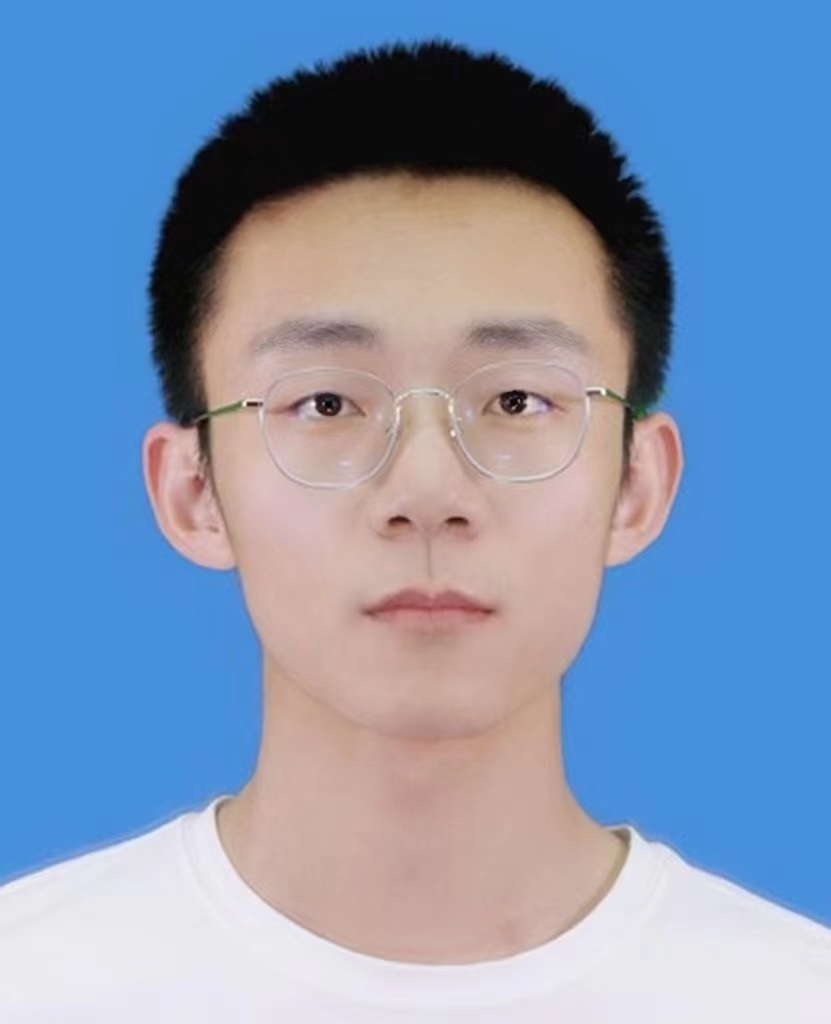}
\end{figure}
\noindent{\bf Yi Song }\quad received the B.E. degree from Beijing Institute of Technology, China, in 2021. He is a master student at the Department of Computer Science and Technology, Tsinghua University. His research interests include natural language generation and dialogue systems.

E-mail: y-song21@mails.tsinghua.edu.cn

\clearpage

\begin{figure}[h]%
\centering
\includegraphics[width=0.3\textwidth]{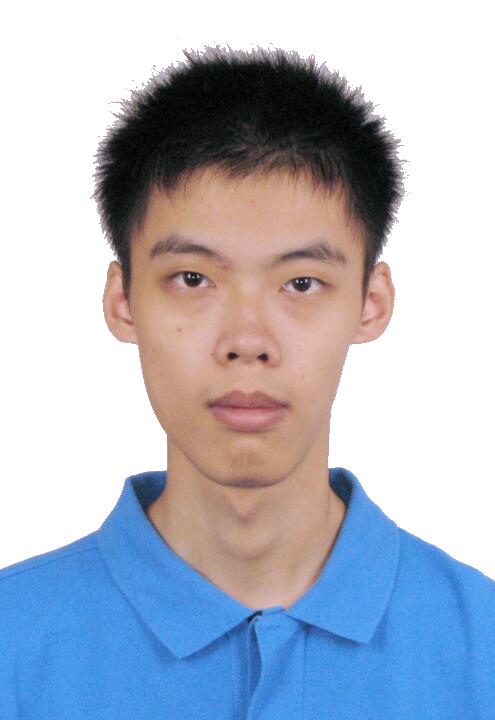}
\end{figure}
\noindent{\bf Pei Ke }\quad received his Ph.D. degree from Tsinghua University, Beijing, China, in 2022. He is currently a postdoctoral researcher at the Department of Computer Science and Technology, Tsinghua University. His research interests include natural language generation, dialogue systems, and sentiment analysis.

E-mail: kepei1106@outlook.com

\begin{figure}[h]%
\centering
\includegraphics[width=0.3\textwidth]{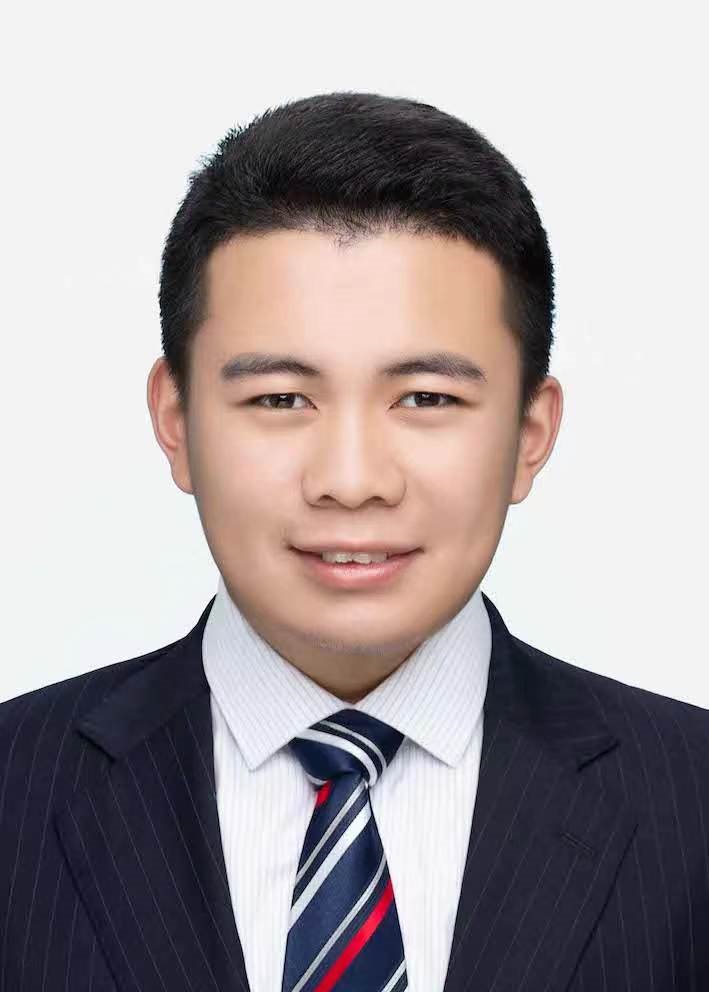}
\end{figure}
\noindent{\bf Chujie Zheng }\quad received the B.S. degree from Tsinghua University, China, in 2020. He is a Ph.D student at the Department of Computer Science and Technology, Tsinghua University. His research interests include natural language generation and dialogue systems.

E-mail: zcj16@tsinghua.org.cn

\clearpage

\begin{figure}[h]%
\centering
\includegraphics[width=0.3\textwidth]{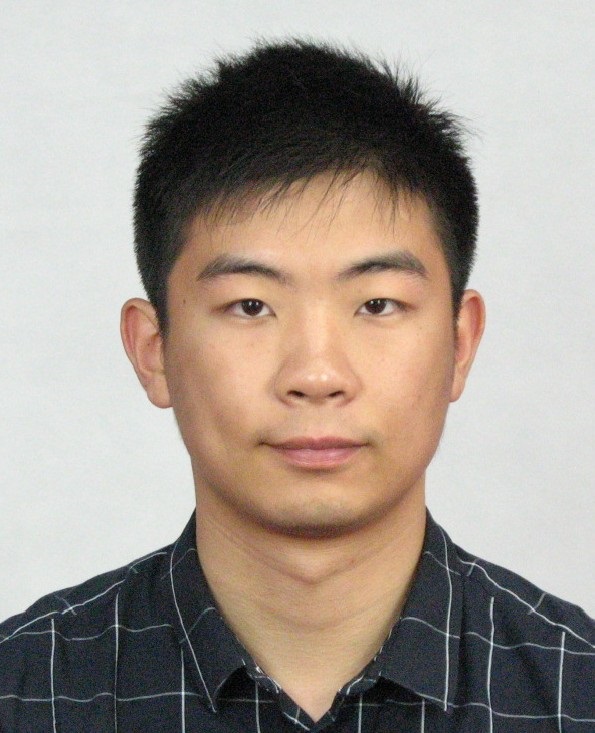}
\end{figure}
\noindent{\bf Zheng Zhang }\quad received his PhD degree from the Department of CS\&T, Tsinghua University in 2021 and the B.E. degree from the same department in 2015. He is now a postdoc researcher at Tsinghua University. His research interests include natural language processing, dialogue systems and text generation.

E-mail: zhangz.goal@gmail.com

\begin{figure}[h]%
\centering
\includegraphics[width=0.3\textwidth]{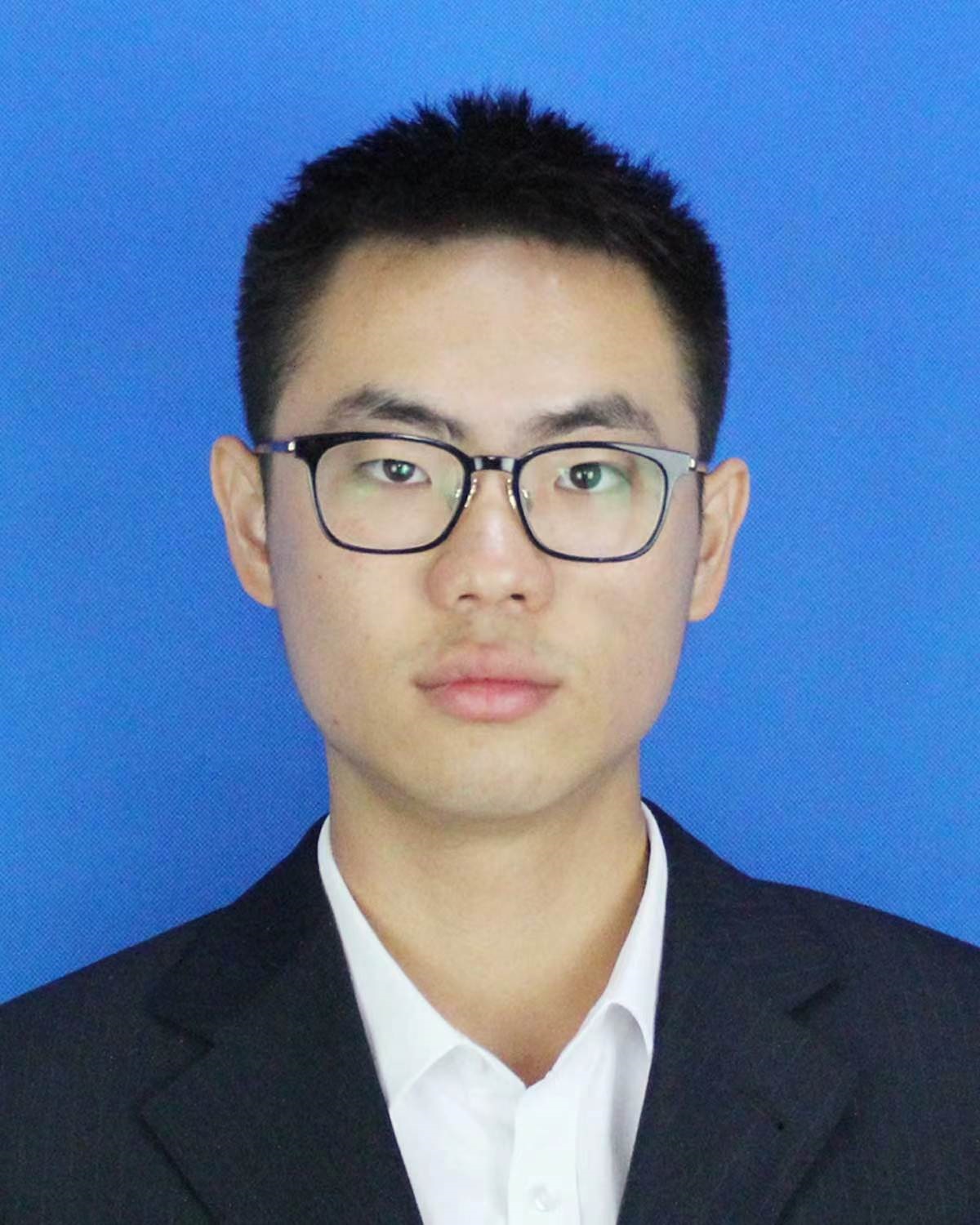}
\end{figure}
\noindent{\bf Jianzhu Yao }\quad is an undergraduate student at the Department of Computer Science and Technology, at Tsinghua University, China. His research interests include natural language generation and dialogue systems.

E-mail: cnyaojz@gmail.com

\clearpage

\begin{figure}[h]%
\centering
\includegraphics[width=0.3\textwidth]{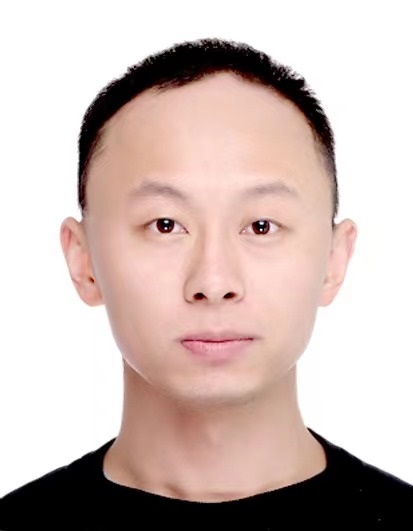}
\end{figure}
\noindent{\bf Lei Liu }\quad received the M.Sc. degree in Computer Science from Central China Normal University, Wuhan, China, in June 2019. He is currently a Ph.D. student in the Graduate Program of Electrical Engineering and Computer Science at York University, Toronto, Canada. His research interests include dialogue systems and natural language generation.

E-mail: lliu@eecs.yorku.ca

\begin{figure}[h]%
\centering
\includegraphics[width=0.3\textwidth]{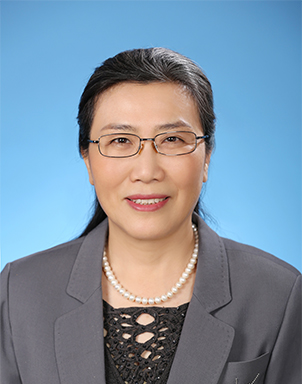}
\end{figure}
\noindent{\bf Xiaoyan Zhu }\quad received the bachelor's degree from the University of Science and Technology Beijing in 1982, master's degree from Kobe University in 1987, and the Ph.D. degree from the Nagoya Institute of Technology, Japan, in 1990. She is currently a Professor with the Department of Computer Science and Technology, Tsinghua University, Beijing, China. Her research interests include intelligent information processing, machine learning, natural language processing, question and answering system and Bioinformatics. She has authored more than 100 peer-reviewed articles in leading international conferences (SIGKDD, IJCAI, AAAI, ACL) and journals (TOIS, Bioinformatics, Genome Biology).

E-mail: zxy-dcs@tsinghua.edu.cn

\clearpage

\begin{figure}[h]%
\centering
\includegraphics[width=0.3\textwidth]{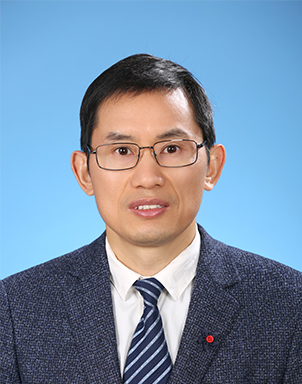}
\end{figure}
\noindent{\bf Minlie Huang }\quad received his Ph.D. degree from Tsinghua University, Beijing, China, in 2006. He is currently an Associate Professor with the Department of Computer Science and Technology, Tsinghua University. His research interests include natural language processing, particularly in dialog systems, reading comprehension, and sentiment analysis. He has published more than 60 papers in premier conferences and journals (ACL, EMNLP, AAAI, IJCAI, WWW, SIGIR, etc.). His work on emotional chatting machines was reported by MIT Technology Review, the Guardian, Nvidia, and many other mass media. He serves as standing reviewer for TACL, area chairs for ACL 2020/2016, EMNLP 2019/2014/2011, and Senior PC members for AAAI 2017-2020 and IJCAI 2017-2020, and reviewers for TASLP, TKDE, TOIS, TPAMI, etc. He is a nominee of ACL 2019 best demo papers, the recipient of IJCAI 2018 distinguished paper award, CCL 2018 best demo award, NLPCC 2015 best paper award, Hanvon Youngth Innovation Award in 2018, and Wuwenjun AI Award in 2019. He was supported by a NSFC key project , several NSFC regular projects, and many IT companies.

E-mail: aihuang@tsinghua.edu.cn

ORCID iD: 0000-0001-7111-1849

\end{document}